\documentclass{llncs}
\usepackage{makeidx}  
\usepackage{amsfonts}
\usepackage{algorithm}
\usepackage{algorithmic}
\usepackage{mathtools}
\usepackage{amssymb}
\usepackage{subfig}
\usepackage{hyperref}
\begin{document}
\frontmatter          
\pagestyle{headings}  
\addtocmark{Hamiltonian Mechanics} 

\mainmatter              
%
\title{Decoding visual stimuli in human brain by using Anatomical Pattern Analysis on fMRI images}


%
\author{Muhammad Yousefnezhad\inst{1} \and Daoqiang Zhang\inst{2}}
%
%
%
\institute{College of Computer Science and Technology, \\Nanjing University of Aeronautics and Astronautics, Nanjing, China\\
\email{\inst{1}myousefnezhad@nuaa.edu.cn, \inst{2}dqzhang@nuaa.edu.cn}}

\maketitle              

\begin{abstract}
A universal unanswered question in neuroscience and machine learning is whether computers can decode the patterns of the human brain. Multi-Voxels Pattern Analysis (MVPA) is a critical tool for addressing this question. However, there are two challenges in the previous MVPA methods, which include decreasing sparsity and noises in the extracted features and increasing the performance of prediction. In overcoming mentioned challenges, this paper proposes Anatomical Pattern Analysis (APA) for decoding visual stimuli in the human brain. This framework develops a novel anatomical feature extraction method and a new imbalance AdaBoost algorithm for binary classification. Further, it utilizes an Error-Correcting Output Codes (ECOC) method for multi-class prediction. APA can automatically detect active regions for each category of the visual stimuli. Moreover, it enables us to combine homogeneous datasets for applying advanced classification. Experimental studies on 4 visual categories (words, consonants, objects and scrambled photos) demonstrate that the proposed approach achieves superior performance to state-of-the-art methods.
\end{abstract}
\section{Introduction}
One of the key challenges in neuroscience is how the human brain activities can be mapped to the different brain tasks. As a conjunction between neuroscience and computer science, Multi-Voxel Pattern Analysis (MVPA) \cite{Norman06} addresses this question by applying machine learning methods on task-based functional Magnetic Resonance Imaging (fMRI) datasets. Analyzing the patterns of visual objects is one of the most interesting topics in MVPA, which can enable us to understand how brain stores and processes the visual stimuli \cite{Haxby14,Osher15}. It can be used for finding novel treatments for mental diseases or even creating a new generation of the user interface in the future.

Technically, there are two challenges in previous studies. The first challenge is decreasing sparsity and noise in preprocessed voxels. Since, most of the previous studies directly utilized voxels for predicting the stimuli, the trained features are mostly sparse, high-dimensional and noisy; and they contain trivial useful information \cite{Haxby14,Osher15,Friston03}. The second challenge is increasing the performance of prediction. Most of the brain decoding problems employed binary classifiers especially by using a one-versus-all strategy \cite{Norman06,Haxby14,Cox03,Mcmenamin15,Mohr15}. In addition, multi-class predictors are even mostly based on the binary classifiers such as the Error-Correcting Output Codes (ECOC) methods \cite{Liu15}. Since task-based fMRI experiments are mostly imbalance, it is so hard to train an effective binary classifier in the brain decoding problems. For instance, consider collected data with 10 same size categories. Since this dataset is imbalance for one-versus-all binary classification, most of the classical algorithms cannot provide acceptable performance \cite{Haxby14,Cox03,Liu09}. 

For facing mentioned problems, this paper proposes Anatomical Pattern Analysis (APA) as a general framework for decoding visual stimuli in the human brain. This framework employs a novel feature extraction method, which uses the brain anatomical regions for generating a normalized view. In practice, this view can enable us to combine homogeneous datasets. The feature extraction method also can automatically detect the active regions for each category of the visual stimuli. Indeed, it can decrease noise and sparsity and increase the performance of the final result. Further, this paper develops a modified version of imbalance AdaBoost algorithm for binary classification. This algorithm uses a supervised random sampling and penalty values, which are calculated by the correlation between different classes, for improving the performance of prediction. This binary classification will be used in a one-versus-all ECOC method as a multi-class approach for classifying the categories of the brain response. 

The rest of this paper is organized as follows: In Section 2, this study briefly reviews some related works. Then, it introduces the proposed method in Section 3. Experimental results are reported in Section 4; and finally, this paper presents conclusion and pointed out some future works in Section 5.
\section{Related Works}
There are three different types of studies for decoding visual stimuli in the human brain. Pioneer studies just focused on the special regions of the human brain, such as the Fusiform Face Area (FFA) or Parahippocampal Place Area (PPA). They only proved that different stimuli can provide different responses in those regions, or found most effective locations based on different stimuli \cite{Haxby14}.

The next group of studies introduced different correlation techniques for understanding similarity or difference between responses to different visual stimuli. Haxby et al. recently showed that different visual stimuli, i.e. human faces, animals, etc., represent different responses in the brain \cite{Haxby14}. Further, Rice et al. proved that not only the mentioned responses are different based on the categories of the stimuli, but also they are correlated based on different properties of the stimuli. They used GIST technique for extracting the properties of stimuli and calculated the correlations between these properties and the brain responses. They separately reported the correlation matrices for different human faces and different objects (houses, chairs, bottles, shoes) \cite{Rice14}. 

The last group of studies proposed the MVPA techniques for predicting the category of visual stimuli. Cox et al. utilized linear and non-linear versions of Support Vector Machine (SVM) algorithm \cite{Cox03}. Norman et al. argued for using SVM and Gaussian Naive Bayes classifiers \cite{Norman06}. Carroll et al. employed the Elastic Net for prediction and interpretation of distributed neural activity with sparse models \cite{Carroll09}. Varoquaux et al. proposed a small-sample brain mapping by using sparse recovery on spatially correlated designs with randomization and clustering. Their method is applied on small sets of brain patterns for distinguishing different categories based on a one-versus-one strategy \cite{Varoquaux12}. McMenamin et al. studied subsystems underlie abstract-category (AC) recognition and priming of objects (e.g., cat, piano) and specific-exemplar (SE) recognition and priming of objects (e.g., a calico cat, a different calico cat, a grand piano, etc.). Technically, they applied SVM on manually selected ROIs in the human brain for generating the visual stimuli predictors \cite{Mcmenamin15}. Mohr et al. compared four different classification methods, i.e. L1/2 regularized SVM, the Elastic Net, and the Graph Net, for predicting different responses in the human brain. They show that L1-regularization can improve classification performance while simultaneously providing highly specific and interpretable discriminative activation patterns \cite{Mohr15}. Osher et al. proposed a network (graph) based approach by using anatomical regions of the human brain for representing and classifying the different visual stimuli responses (faces, objects, bodies, scenes) \cite{Osher15}. 
\begin{figure}[!t]
	\centering
	\includegraphics[width=4.5in]{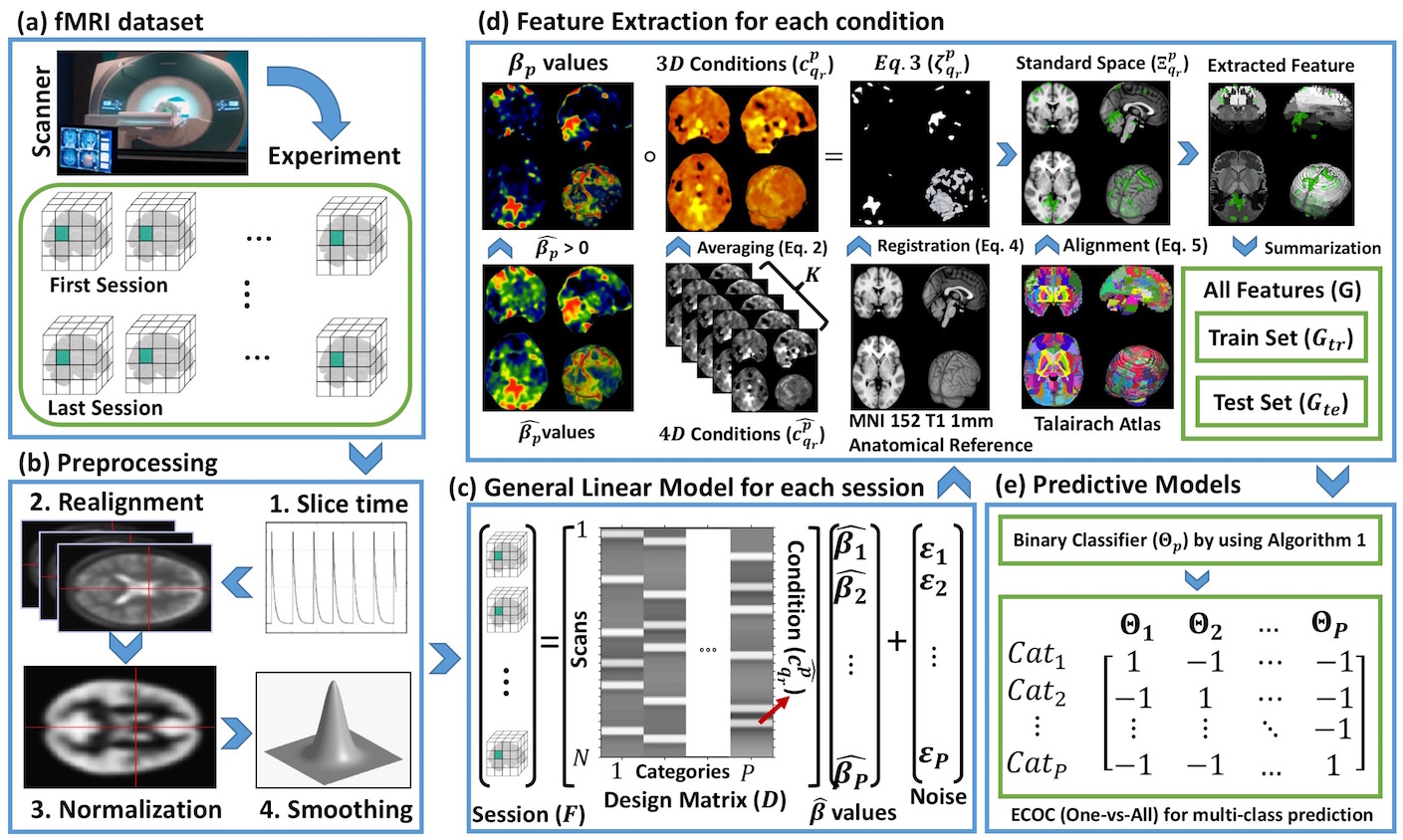}
	\vskip -0.1 in
	\caption{Anatomical Pattern Analysis (APA) framework}
	\vskip -0.27in
\end{figure}
\section{The Proposed Method}
Blood Oxygen Level Dependent (BOLD) signals are used in fMRI techniques for representing the neural activates. Based on hyperalignment problem in the brain decoding \cite{Haxby14}, quantity values of the BOLD signals in the same experiment for the two subjects are usually different. Therefore, MVPA techniques use the correlation between different voxels as the pattern of the brain response \cite{Osher15,Friston03}. As depicted in Figure 1, each fMRI experiment includes a set of sessions (time series of 3D images), which can be captured by different subjects or just repeating the imaging procedure with a unique subject. Technically, each session can be partitioned into a set of visual stimuli categories. Indeed, an independent category denotes a set of homogeneous conditions, which are generated by using the same type of photos as the visual stimuli. For instance, if a subject watches 6 photos of cats and 5 photos of houses during a unique session, this 4D image includes 2 different categories and 11 conditions.
\subsection{Feature Extraction}
Consider $F \in \mathbb{R}^{N\times X\times Y \times Z} = \{$number of scans $(N) \times$ 3D images$\}$ for each session of the experiment. $F$ can be written as a general linear model: $F = D\beta+\varepsilon$, where $D = \{$number of scans (N) $\times$  $P$ categories (regressors)$\}$ denotes the design matrix; $\varepsilon$ is the noise (error of estimation); and also $\beta = \{$number of categories $(P) \times$ 3D images$\}$ denotes the set of correlations between voxels for the categories of the session. Design matrix can be calculated by convolution $(D(t) = (S*H)(t))$ of onsets (or time series $S(t)$) and the Hemodynamic Response Function (HRF) \cite{Friston03}. This paper uses Generalized Least Squares (GLS) approach for estimating optimized solution $(\hat{\beta}={({D}^{\intercal}{V}^{-1}D)}^{-1}{D}^{\intercal}{V}^{-1}F)$, where $V$ is the covariance matrix of the noise ($Var(\varepsilon)=V{\sigma}^{2}\ne\mathbb{I}{\sigma}^{2}$) \cite{Friston03,Haxby14}. Now, this paper defines the positive correlation $\beta = \hat{\beta} > 0 = \{\hat{{\beta}_{1}} > 0, \hat{{\beta}_{2}} > 0, \dots, \hat{{\beta}_{P}} > 0\}= \{{\beta}_{1}, {\beta}_{2}, \dots, {\beta}_{P}\}$ for all categories as the active regions, where $\hat{\beta}$ denotes the estimated correlation, $\hat{\beta}_{p}$ and ${\beta}_{p}$ are the correlation and \emph{positive} correlation for the $p$-th category, respectively. Moreover, the data $F$ is partitioned based on the conditions of the design matrix as follows: 
\begin{equation}
\hat{C} = \{\hat{c}_{1}^{1}, \hat{c}_{2}^{1}, \dots, \hat{c}_{Q_1}^{1},\hat{c}_{1}^{2}, \hat{c}_{2}^{2}, \dots, \hat{c}_{Q_2}^{2},\dots,\hat{c}_{1}^{p}, \hat{c}_{2}^{P}, \dots, \hat{c}_{Q_R}^{P}\}
\label{EqConditions}
\end{equation}
where $\hat{C}$ denotes the set of all conditions in each session, $P$ and $Q_r$ are respectively the number of categories in each session and the number of conditions in each category. Further, $\hat{c}^{p}_{q_r}$ = \{number of scans $({K}^{p}_{q_r})$ $\times$ 3D images\} denotes the 4D images for the $p$-th category and ${q_r}$-th condition in the design matrix. Now, this paper defines the sum of all images in a condition as follows:
\begin{equation}
C^p_{q_r} = \sum_{K}\hat{c}^{p}_{q_r}=\sum_{k=1}^{{K}^{p}_{q_r}}\hat{c}^{p}_{q_r}[k,:,:,:]
\label{EqRawFeatures}
\end{equation}
where ${c}^{p}_{q_r}[k,:,:,:]$ denotes all voxels in the $k$-th scan of $q_r$-th condition of $p$-th category; also ${K}^{p}_{q_r}$ is the number of scans in the given condition. ${\zeta}^{p}_{q_r}$ matrix is denoted for applying the correlation of voxels on the response of each condition as follows: 
\begin{equation}
{\zeta}^{p}_{q_r} = \beta_p \circ C_{q_r}^{p} = \{\forall [x,y,z] \in C^{p}_{q_r} \implies {({\zeta}}^{p}_{q_r})_{[x,y,z]} = {(\beta_p)}_{[x,y,z]} \times {(C_{q_r}^p)}_{[x,y,z]}\}  
\label{Eq:RawFeatures}
\end{equation}
where $\circ$ denotes Hadamard product; and $(C_{q_r}^p)_{[x,y,z]}$ is the $[x,y,z]$-th voxel of the $q_r$-th condition of $p$-th category; and also, ${(\beta_p)}_{[x,y,z]}$ is the $[x,y,z]$-th voxel of the correlation matrix ($\beta$ values) of the $p$-th category. 

Since mapping 4D fMRI images to standard space decreases the performance of final results, most of the previous studies use the original images instead of the standard version. By considering 3D image $\zeta_{q_r}^p$ for each condition, this paper enables to map brain activities to a standard space. This mapping can provide normalized view for combing homogeneous datasets. For registering $\zeta_{q_r}^p$ to standard space, this paper utilizes the FLIRT algorithm \cite{Jenkinson02}, which minimizes the following cost function:
\begin{equation}
T^* = {argmin}_{T \in S_T} (NMI(Ref, \Xi_{q_r}^p))
\label{EqImageReg}
\end{equation}
where $Ref$ denotes the reference image, $S_T$ is the space of allowable transformations, the function $NMI$ denotes the Normalized Mutual Information between two images, $\Xi_{q_r}^p = T({\zeta}_{q_r}^{p})$ is the condition after registration ($T$ denotes the transformation function) \cite{Jenkinson02}. The performance of (\ref{EqImageReg}) will be analyzed in Section 4. Now, consider $Atlas = \{A_1,A_2,\dots,A_L\}$, where ${\cap}_{l=1}^{L}\{A_l\}=\emptyset$, ${\cup}_{l=1}^{L}\{A_l\}=A$ and $A_l$ denotes the set of indexes of voxels for the $l$-th region. The extracted feature for $l$-th region of $q_r$-th condition of $p$-th category is calculated as follows, where ${a}_{v} = [x_v,y_v,z_v]$ denotes the index of $v$-th voxel of $l$-th atlas region; and $A_l$ is the set of indexes of voxels in the $l$-th region. 
\begin{equation}
\forall {a}_{v}=[x_v,y_v,z_v] \in A_l \implies {\Gamma}_{q_r}^p(l)=\frac{1}{\mid A_l \mid}\sum_{v = 1}^{\mid A_l \mid}{({\Xi}_{q_r}^{p})[a_v]}=\frac{1}{\mid A_l \mid}\sum_{v = 1}^{\mid A_l \mid}{({\Xi}_{q_r}^{p})[x_v,y_v,z_v]}
\label{FeatureExtraction}
\end{equation}
\subsection{Classification Algorithm}
This paper randomly partitions the extracted features $G = \{[{\Gamma}_{1}^{1}(1) \dots {\Gamma}_{1}^{1}(L)],\dots,$ $[{\Gamma}_{Q_1}^{1}(1) \dots {\Gamma}_{Q_1}^{1}(L)], \dots,[{\Gamma}_{Q_R}^{P}(1) \dots {\Gamma}_{Q_R}^{P}(L)] \}$ to the train set $(G_{tr})$ and the test set $(G_{te})$. As a new branch of AdaBoost algorithm, Algorithm 1 employs $G_{tr}$ for training binary classification. Then, $G_{te}$ is utilized for estimating the performance of the classifier. As mentioned before, training binary classification for fMRI analysis is mostly imbalance, especially by using a one-versus-all strategy. As a result, the number of samples in one of these binary classes is smaller than the other class. This paper also exploits this concept. Indeed, Algorithm 1 firstly partitions the train data $({G}_{tr})$ to small $({G}_{tr}^{S})$ and large $({G}_{tr}^{L})$ classes (groups) based on the class labels $({I}_{tr} \in \{+1,-1\})$. Then, it calculates the scale $(J)$ of existed elements between two classes; and employs this scale as the number of the ensemble iteration $(J+1)$. Here, $Int()$ denotes the floor function. In the next step, the large class is randomly partitioned to $J$ parts. Now, train data $(G_j)$ for each iteration is generated by all instances of the small class $({G}_{tr}^{S})$, one of the partitioned parts of the large class $({G}_{tr}^{L}(j))$ and the instances of the previous iteration $(\bar{G_j})$, which cannot truly be trained. In this algorithm, $corr()$ function denotes the Pearson correlation $(corr(A,B) = cov(A,B)/\sigma_A \sigma_B)$; and $W_j \in [0, 1]$ is the train weight (penalty values), which is considered for the large class. Further, $Classifier()$ denotes any kind of weighted classification algorithm. This paper uses a simple classical decision tree as the individual classification algorithm ($\theta_j$) \cite{Liu09}.\\

Generally, there are two techniques for applying multi-class classification. The first approach directly creates the classification model such as multi-class support vector machine \cite{Cox03} or neural network \cite{Norman06}. In contrast, (indirect) decomposition design uses an array of binary classifiers for solving the multi-class problems. As one of the classical indirect methods, Error-Correcting Output Codes (ECOC) includes three components, i.e. base algorithm, encoding and decoding procedures \cite{Liu15}. As the based algorithm in the ECOC, this paper employs Algorithm 1 for generating the binary classifiers ($\Theta_p$). Further, it uses a one-versus-all encoding strategy for training the ECOC method, where an independent category of the visual stimuli is compared with the rest of categories (see Figure 1.e). Indeed, the number of classifiers in this strategy is exactly equal to the number of categories. This method also assigns the brain response to the category with closest hamming distance in decoding stage.
\begin{algorithm}[!t]
	\caption{The proposed binary classification algorithm}
	\label{alg:AdaBoost}
	\begin{algorithmic}
		\STATE {\bfseries Input:} Data set ${G}_{tr}:$ is train set, ${I}_{tr}:$ denotes real class labels of the train sets,\\
		\STATE {\bfseries Output:} Classifier $E$,
		\STATE {\bfseries Method:}\\
		\quad1. Partition ${G}_{tr} = \{{G}_{tr}^{S}, {G}_{tr}^{L}\}$, where ${G}_{tr}^{S}$, ${G}_{tr}^{L}$ are Small and Large classes.\\
		\quad2. Calculate $J = Int(\mid {G}_{tr}^{S} \mid / \mid {G}_{tr}^{L} \mid)$ based on number of elements in classes.\\
		\quad3. Randomly sample the ${G}_{tr}^{L} = \{{G}_{tr}^{L}(1), \dots, {G}_{tr}^{L}(J) \}$.\\
		\quad4. By considering $\bar{G_1}=\bar{I_1}=\emptyset$, generating $j = 1, \dots, J+1$ classifiers:\\
		\quad5. Construct $G_j = \{{G}_{tr}^{S},{G}_{tr}^{L}(j),\bar{G_j}\}$ and $I_j = \{{I}_{tr}^{S},{I}_{tr}^{L}(j),\bar{I_j}\}$
		\[ \text{6. Calculate } W_j = {\{ w_j \}}_{\mid G_j \mid } =
		\begin{cases}
		1   & \quad \text{for instances of } {G}_{tr}^{S} \text{ or } \bar{G_j}\\
		1 - \mid corr({G}_{tr}^{S}, {G}_{tr}^{L}) \mid & \quad \text{for instances of } {G}_{tr}^{L}(j)\\
		\end{cases}\]\\
		\quad7.Train $\theta_j = Classifier(G_j, I_j, W_j)$.\\
		\quad8. Construct $\bar{{G}}_{j+1}$, $\bar{{I}}_{j+1}$ as the set of instances cannot truly trained in $\theta_j$.\\
		\quad9. \textbf{If} $(j \leq J + 1)$: go to line $5$; \textbf{Else:} return $\Theta_p = \{\theta_1, \dots, \theta_{J+1}\} $ as final classifier.\\ 
	\end{algorithmic}
\end{algorithm}
\section{Experiments}
\subsection{Extracted Features Analysis}
This paper employs two datasets, shared by \url{openfmri.org}, for running empirical studies. As the first dataset, `Visual Object Recognition' (DS105) includes 71 sessions (6 subjects). It also contains 8 categories of visual stimuli, i.e. gray-scale images of faces, houses, cats, bottles, scissors, shoes, chairs, and scrambled (nonsense) photos. This dataset is analyzed in high-level visual stimuli as the binary predictor, by considering all categories except scrambled photos as objects, and low-level visual stimuli in the multi-class prediction. Please see \cite{Haxby14,Cox03} for more information. As the second dataset, `Word and Object Processing' (DS107) includes 98 sessions (49 subjects). It contains 4 categories of visual stimuli, i.e. words, objects, scrambles, consonants. Please see \cite{Duncan09} for more information. These datasets are preprocessed by SPM 12 (\url{www.fil.ion.ucl.ac.uk/spm/}), i.e. slice timing, realignment, normalization, smoothing. Then, the beta values are calculated for each session. This paper employs the \textit{MNI 152 T1 1mm} (see Figure 1.d) as the reference image ($Ref$) in Eq. (4) for registering the extracted conditions ($\zeta$) to the standard space ($\Xi$). In addition, this paper uses \textit{Talairach} Atlas (contains $L=1105$ regions) in Eq. (5) for extracting features (See Figure 1.d).

Figures 2.a-c demonstrate examples of brain responses to different stimuli, i.e. (a) word, (b) object, and (c) scramble. 
Here, gray parts show the anatomical atlas, the colored parts (red, yellow and green) define the functional activities, and also the red rectangles illustrate the error areas after registration. Indeed, these errors can be formulated as the nonzero areas in the brain image which are located in the zero area of the anatomical atlas (the area without region number). The performance of objective function \eqref{EqImageReg} on DS105, and DS107 data sets is analyzed in Figure 2.d by using different distance metrics, i.e. Woods function (W), Correlation Ratio (CR), Joint Entropy (JE), Mutual Information (MI), and Normalized Mutual Information (NMI) \cite{Jenkinson02}. As depicted in this figure, the NMI generated better results in comparison with other metrics. 

Figure 3.a and c illustrate the correlation matrix of the DS105 and DS107 at the voxel level, respectively. Similarly, Figure 3.b and d show the correlation matrix the DS105 and DS107 in the feature level, respectively. Since, brain responses are sparse, high-dimensional and noisy at voxel level, it is so hard to discriminate between different categories in Figure 2.a and c. By contrast, Figure 2.b and d provide distinctive representation when the proposed method used the correlated patterns in each anatomical regions as the extracted features.
\begin{figure}[!t]
	\centering
	\includegraphics[width=0.60\textwidth,height=0.55\linewidth]{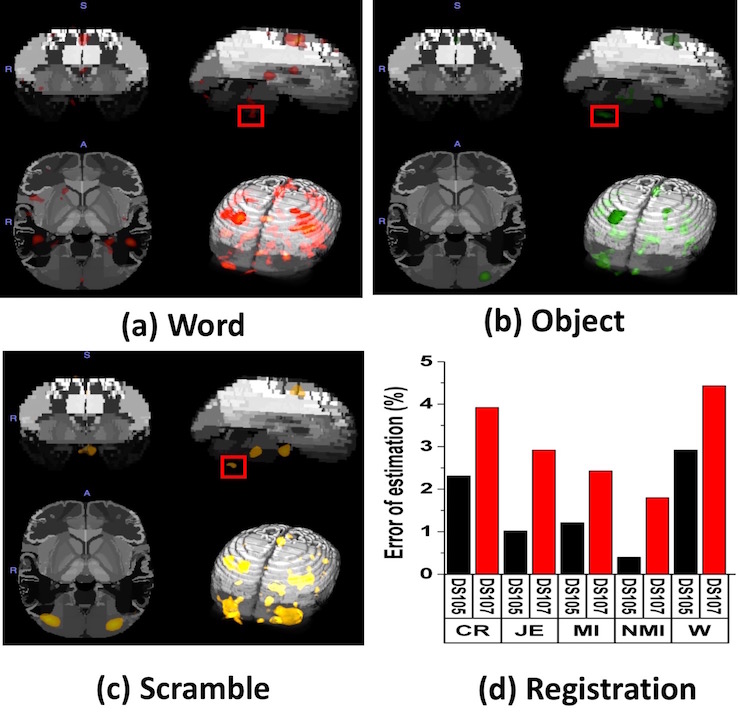}
	\vskip -0.1 in
	\caption{Extracted features based on different stimuli, i.e. (a) word, (b) object, and (c) scramble. (d) The effect of different objective functions in (\ref{EqImageReg}) on the error of registration.\\}
	\vskip -0.3in
\end{figure}
\begin{figure}[!t]
	\begin{center}
		\begin{minipage}{0.75\linewidth}
			\includegraphics[width=0.98\textwidth]{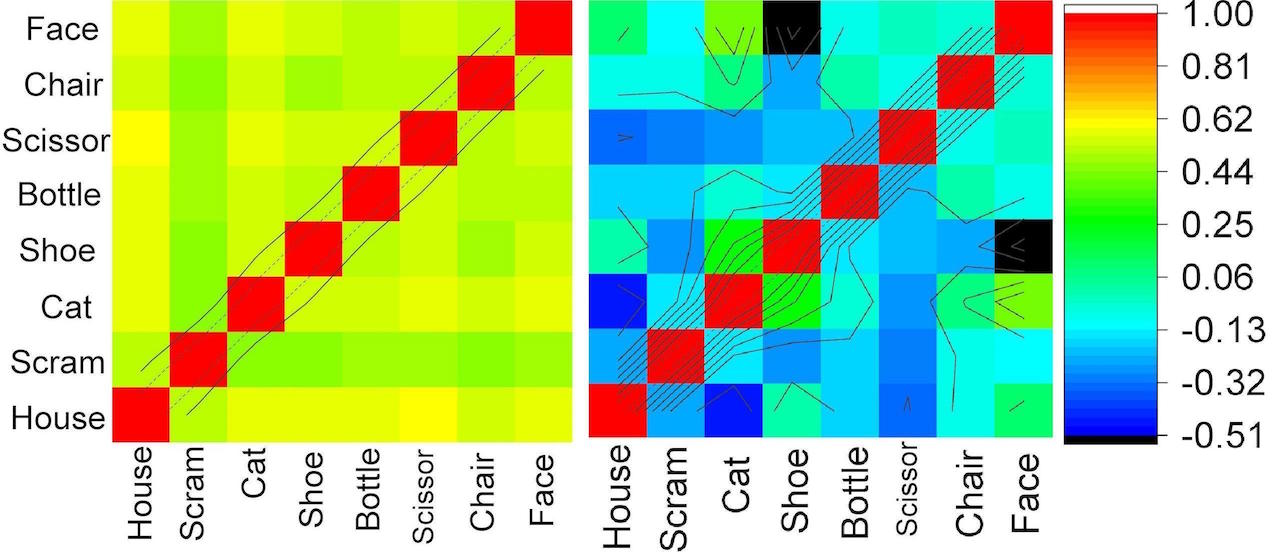}\\
			\centering (a)  \qquad\qquad\qquad\qquad (b)\\ 
		\end{minipage}
		\begin{minipage}{0.75\linewidth}
			\includegraphics[width=0.98\textwidth]{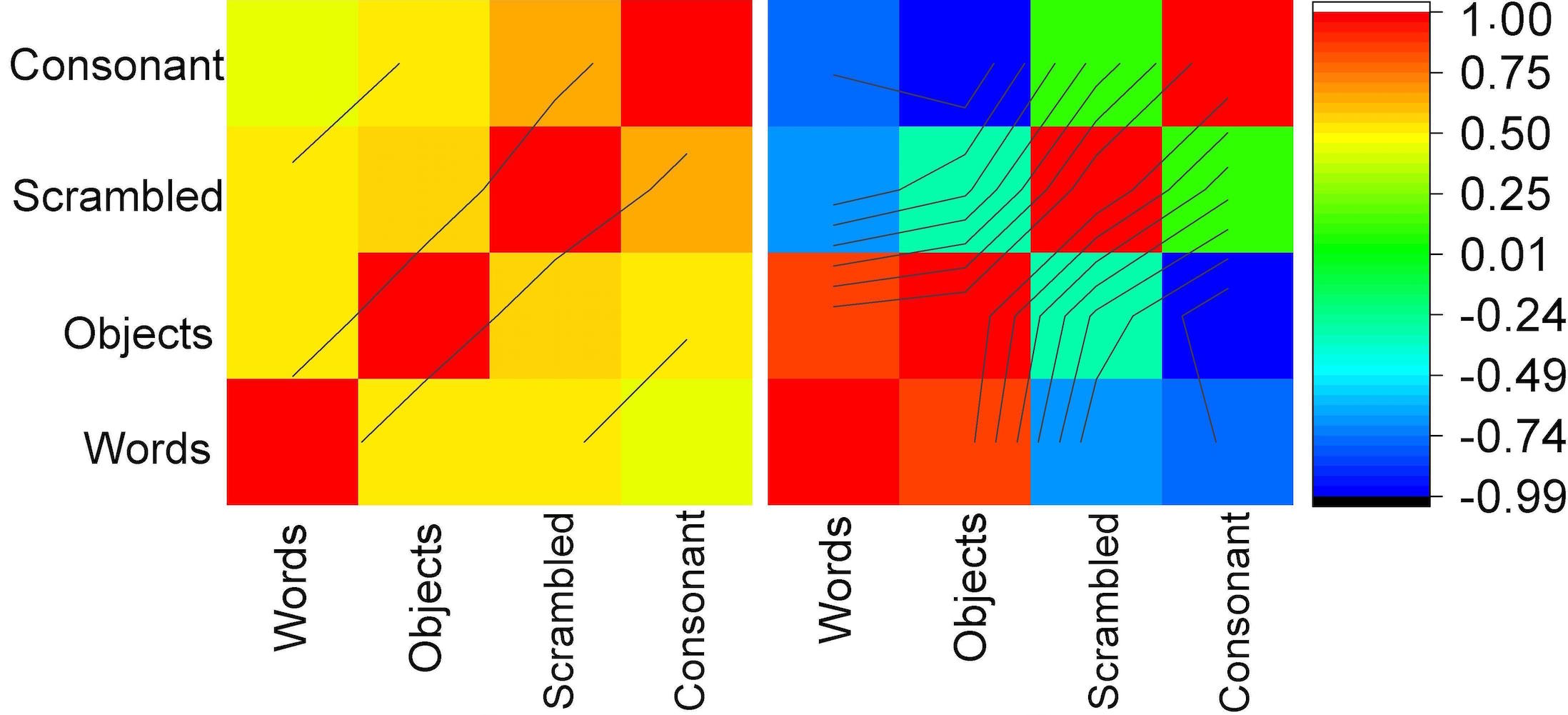}\\
			\centering (c) \qquad\qquad\qquad\qquad (d) 
		\end{minipage}		
		\caption{The correlation matrices: (a) raw voxels and (b) extracted features of the DS105 dataset, (c) raw voxels and (d) extracted features of the DS107 dataset.}
		\label{SmoothedDM}
	\end{center}
	\vskip -0.3in
\end{figure}
\subsection{Classification Analysis}
The performance of our framework is compared with state-of-the-art methods, i.e. Cox \& Savoy \cite{Cox03}, McMenamin et al. \cite{Mcmenamin15}, Mohr et al. \cite{Mohr15}, and Osher et al. \cite{Osher15}, by using leave-one-out cross validation in the subject level. Further, all of algorithms are implemented in the MATLAB R2016a (9.0) by authors in order to generate experimental results. Tables 1 and 2 respectively illustrate the classification Accuracy and Area Under the ROC Curve (AUC) for the binary predictors based on the category of the visual stimuli. All visual stimuli in the dataset DS105 except scrambled photos are considered as the object category for generating these experimental results. As depicted in the Tables 1 and 2, the proposed algorithm has achieved better performance in comparison with other methods because it provided a better representation of neural activities by exploiting the anatomical structure of the human brain. Table 3 illustrates the classification accuracy for multi-class predictors. In this table, `DS105' includes 8 different categories (P=8 classes) and `DS107' contains 4 categories of the visual stimuli. As another 4 categories dataset, `ALL' is generated by considering all visual stimuli in the dataset DS105 except scrambled photos as object category and combining them with the dataset DS107. In this dataset, the accuracy of the proposed method is improved by combining two datasets, whereas, the performances of other methods are significantly decreased. As mentioned before, it is the standard space registration problem in the 4D images. In addition, our framework employs the extracted features from the brain structural regions instead of using all or a subgroup of voxels, which can increase the performance of the predictive models by decreasing noise and sparsity. 
\begin{table*}[!t]
	\centering
	\caption{Accuracy of binary predictors  }
	\vskip -0.15in
	\label{tbl: BinaryAccuracy}
	\begin{center}
		\begin{small}
			\begin{tabular}{lccccc}
				\hline
				Data Sets & Cox \& Savoy & McMenamin et al. & Mohr el al. & Osher et al. & Binary-APA \\
				\hline
				DS105-Objects & 71.65$\pm$0.97 & 83.06$\pm$0.36 & 85.29$\pm$0.49 & 90.82$\pm$1.23 & \textbf{98.37$\pm$0.16}  \\
				DS107-Words & 69.89$\pm$1.02 & 89.62$\pm$0.52 & 81.14$\pm$0.91 & 94.21$\pm$0.83 & \textbf{97.67$\pm$0.12}  \\
				DS107-Consonants & 67.84$\pm$0.82 & 87.82$\pm$0.37 & 79.69$\pm$0.69 & 95.54$\pm$0.99 & \textbf{98.73$\pm$0.06}\\
				DS107-Objects & 65.32$\pm$1.67 & 84.22$\pm$0.44 & 75.32$\pm$0.41 & \textbf{95.62$\pm$0.83} & {95.06$\pm$0.11}\\
				DS107-Scramble & 67.96$\pm$0.87 & 86.19$\pm$0.26 & 78.45$\pm$0.62 & 93.1$\pm$0.78 & \textbf{96.71$\pm$0.18}\\
				\hline
			\end{tabular}
		\end{small}
	\end{center}
	\vskip -0.35in
\end{table*}
\begin{table*}[!t]
	\centering
	\caption{Area Under the ROC Curve (AUC) of binary predictors  }
	\vskip -0.15in
	\label{tbl: BinaryAUC}
	\begin{center}
		\begin{small}
			\begin{tabular}{lccccc}
				\hline
				Data Sets & Cox \& Savoy & McMenamin et al. & Mohr el al. & Osher et al. & Binary-APA \\
				\hline
				DS105-Objects & 68.37$\pm$1.01 & 82.22$\pm$0.42 & 80.91$\pm$0.21 & 88.54$\pm$0.71 & \textbf{96.25$\pm$0.92}  \\
				DS107-Words & 67.76$\pm$0.91 & 86.35$\pm$0.39 & 78.23$\pm$0.57 & 93.61$\pm$0.62 & \textbf{97.02$\pm$0.2}  \\
				DS107-Consonants & 63.84$\pm$1.45  & 85.63$\pm$0.61 & 77.41$\pm$0.92 & 94.54$\pm$0.31 & \textbf{96.92$\pm$0.14}\\
				DS107-Objects & 63.17$\pm$0.59  & 81.54$\pm$0.92 & 73.92$\pm$0.28 & 94.23$\pm$0.94 & \textbf{95.17$\pm$0.03}\\
				DS107-Scramble & 66.73$\pm$0.92 & 85.79$\pm$0.42 & 76.14$\pm$0.47 & 92.23$\pm$0.38 & \textbf{96.08$\pm$0.1}\\
				\hline
			\end{tabular}
		\end{small}
	\end{center}
	\vskip -0.35in
\end{table*}
\begin{table*}[!t]
	\centering
	\caption{Accuracy of multi-class predictors}
	\vskip -0.15in
	\label{tbl: MultiACC}
	\begin{center}
		\begin{small}
			\begin{tabular}{lccccc}
				\hline
				Data Sets & Cox \& Savoy & McMenamin et al. & Mohr el al. & Osher et al. &  Multi-APA \\
				\hline
				DS105 (P=8) & 18.03$\pm$4.07 & 38.34$\pm$3.21 & 29.14$\pm$2.25 & 50.61$\pm$4.83 & \textbf{57.93$\pm$2.1} \\
				DS107 (P=4) & 38.01$\pm$2.56 & 71.55$\pm$2.79 & 64.71$\pm$3.14 & 89.69$\pm$2.32 & \textbf{94.21$\pm$2.41}\\
				ALL     (P=4) & 32.93$\pm$2.29 & 68.35$\pm$3.07 & 63.16$\pm$4 & 80.36$\pm$3.04 & \textbf{95.67$\pm$1.25}\\
				\hline
			\end{tabular}
		\end{small}
	\end{center}
	\vskip -0.3in
\end{table*}

\section{Conclusion}
This paper proposes Anatomical Pattern Analysis (APA) framework for decoding visual stimuli in the human brain. This framework uses an anatomical feature extraction method, which provides a normalized representation for combining homogeneous datasets. Further, a new binary imbalance AdaBoost algorithm is introduced. It can increase the performance of prediction by exploiting a supervised random sampling and the correlation between classes. In addition, this algorithm is utilized in an Error-Correcting Output Codes (ECOC) method for multi-class prediction of the brain responses. Empirical studies on 4 visual categories clearly show the superiority of our proposed method in comparison with the voxel-based approaches. In future, we plan to apply the proposed method to different brain tasks such as low-level visual stimuli, emotion and etc. 
\vskip -0.1in
\section*{Acknowledgment}
We thank the anonymous reviewers for comments. This work was supported in part by the National Natural Science Foundation of China (61422204 and 61473149), Jiangsu Natural Science Foundation for Distinguished Young Scholar (BK20130034) and NUAA Fundamental Research Funds (NE2013105).

\clearpage
\addtocmark[2]{Author Index} 
\renewcommand{\indexname}{Author Index}
\printindex
\clearpage
\addtocmark[2]{Subject Index} 
\markboth{Subject Index}{Subject Index}
\end{document}